\documentclass{article}

\usepackage{arxiv}

\usepackage[utf8]{inputenc} 
\usepackage[T1]{fontenc}    
\usepackage{hyperref}       
\usepackage{url}            
\usepackage{booktabs}       
\usepackage{amsfonts}       
\usepackage{nicefrac}       
\usepackage{microtype}      
\usepackage{lipsum}
\usepackage{amsmath,amssymb,amsfonts, algorithm2e}
\usepackage{algorithmic}
\usepackage{graphicx}
\usepackage{textcomp}
\usepackage{xcolor}
\newcommand{\etal}{\textit{et al.}}
\usepackage{multirow}

\title{GreenPCO: An Unsupervised Lightweight \\
Point Cloud Odometry Method}

\author{
  Pranav Kadam \\
  Media Communications Lab \\
  University of Southern California\\
  Los Angeles, CA, USA \\
  \texttt{pranavka@usc.edu} \\
  
  \And
 Min Zhang \\
  Media Communications Lab\\
  University of Southern California\\
  Los Angeles, CA, USA \\
  \texttt{zhan980@usc.edu} \\
  
    \And
 Jiahao Gu \\
  Media Communications Lab\\
  University of Southern California\\
  Los Angeles, CA, USA \\
  \texttt{jiahaogu@usc.edu} \\
  
   \And
 Shan Liu\thanks{This work was supported by a gift grant from Tencent.} \\
  Tencent Media Lab\\
  Tencent America\\
  Palo Alto, CA, USA \\
  \texttt{shanl@tencent.com} \\
  
  \And
 C.-C. Jay Kuo \\
  Media Communications Lab\\
  University of Southern California\\
  Los Angeles, CA, USA \\
  \texttt{cckuo@sipi.usc.edu} \\
}

\begin{document}
\maketitle

\begin{abstract}
Visual odometry aims to track the incremental motion of an object using
the information captured by visual sensors. In this work, we study the
point cloud odometry problem, where only the point cloud scans obtained
by the LiDAR (Light Detection And Ranging) are used to estimate object's
motion trajectory. A lightweight point cloud odometry solution is
proposed and named the green point cloud odometry (GreenPCO) method.  GreenPCO
is an unsupervised learning method that predicts object motion by
matching features of consecutive point cloud scans. It consists of three
steps. First, a geometry-aware point sampling scheme is used to select
discriminant points from the large point cloud. Second, the view is
partitioned into four regions surrounding the object, and the PointHop++
method is used to extract point features. Third, point correspondences
are established to estimate object motion between two consecutive scans.
Experiments on the KITTI dataset are conducted to demonstrate the
effectiveness of the GreenPCO method.  It is observed that GreenPCO outperforms
benchmarking deep learning methods in accuracy while it has a
significantly smaller model size and less training time. 
\end{abstract}


\section{Introduction}\label{sec:introduction}

Odometry is an object localization technique that estimates the position
change of an object by sensing changes in the surrounding environment over
time. It finds a range of applications such as the navigation of mobile
robots and autonomous vehicles. Odometry is also responsible for the
localization task in an SLAM (Simultaneous Localization And Mapping)
system.  
When only the visual information is exploited, it is called visual
odometry.  Several visual odometry solutions that use monocular,
monochrome and stereo vision have been proposed and successfully
deployed.  With the growing popularity of 3D point clouds, point cloud
scans obtained by range sensors such as LiDAR (Light Detection And
Ranging) are recently used in odometry, which is known as the point
cloud odometry (PCO). 

It is worthwhile to mention that multi-modal data from different sensors
such as motion sensors (e.g, wheel encoders), inertial measurement unit
(IMU) and the geo positioning system (GPS) are often used jointly to
boost localization accuracy since different sensors provide
complementary information. However, the accuracy of multi-sensor
odometry is still built upon that of each individual sensor. Generally,
since techniques of improving PCO accuracy and performance enhancement
via multi-modal sensor fusion are of different nature, they can be
treated separately. 

There is a recent trend in the design of deep neural networks for point
cloud odometry. They replace the matching of traditional handcrafted
features with the end-to-end optimized network models. On one hand, deep
learning (DL) is promising in handling several long standing scan
matching problems, e.g., matching in presence of noisy data and
outliers, matching in featureless environments, etc. On the other hand,
the generalization of deep learning models between different datasets is
poor.  Different datasets are needed for different applications.  For
example, datasets for indoor robot navigation and self driving vehicles
have to be collected separately.  Furthermore, these methods are based
on supervised learning that demands the ground truth transformation
parameters in network training, which contradicts classical methods that
do not need any ground truth and are purely based on local geometric
properties of point clouds. 

In this work, we focus on the PCO problem by proposing a lightweight PCO
solution called the green point cloud odometry (GreenPCO) method.
GreenPCO is an unsupervised learning method that predicts object motion
by matching features of two consecutive point cloud scans.  In the
training stage, a small number of point cloud scans from the training
dataset are used to learn model parameters of GreenPCO in the
feedforward one-pass manner.  In the inference stage, GreenPCO finds the
vehicle trajectory online by incrementally predicting the motion between
two consecutive point cloud scans.  We call it green due to its smaller
model size and significantly less training time as compared with other
learning-based PCO methods. 

Our research is inspired by a new unsupervised point cloud registration
method called R-PointHop \cite{kadam2021r}. The problem is to find the rigid transformation of a point cloud set that is viewed in different
translated and rotated coordinates.  The point cloud registration
technique can be extended to point cloud odometry in principle. Yet, to
tailor it to the odometry task, several modifications to R-PointHop are
needed.  

GreenPCO consists of three steps.  First, a subset of discriminant points is
sampled from the original point cloud using a geometry-aware point
sampling scheme, and sampled points are divided into four mutually
disjoint groups with view-based partitioning. Second, point features are
derived using the learned GreenPCO model and point correspondences are built
using feature matching.  Third, the motion between the two scans is
predicted using the singular value decomposition (SVD) of the matrix of
point correspondences.  This process repeats for every two consecutive
scans in time.  Experiments on the KITTI dataset \cite{Geiger2012CVPR}
are conducted to demonstrate the effectiveness of the GreenPCO method. GreenPCO
outperforms benchmarking deep learning (DL) methods in accuracy while it
has a significantly smaller model size and less training time. 

\section{Related Work}\label{sec:review}

\subsection{Model free methods} \label{subsec:model-free}

The state-of-the-art point cloud odometry method is LOAM
\cite{zhang2014loam}.  It is a combination of two algorithms running in
parallel - one for LiDAR odometry and the other for point cloud
registration.  LOAM adopts a pipeline which includes scan matching,
motion estimation and mapping.  V-LOAM (Visual LOAM)
\cite{zhang2015visual} combines LiDAR and image data by exploiting the
advantages of multi-modality sensors.  LoL \cite{rozenberszki2020lol}
integrates LOAM and segment matching with respect to an offline map to
compensate for odometry drift. ORB-SLAM \cite{mur2015orb} is a SLAM
system that leverages monocular vision.  A new camera calibration
technique that improves the visual odometry estimates of several high
performance methods was proposed in \cite{Geiger2012CVPR} for the KITTI
dataset \cite{cvivsic2021recalibrating}.  There are also a few
geometry-based point cloud odometry methods that perform scan matching
using the iterative closest point (ICP) algorithm \cite{besl1992method}
or its variants \cite{rusinkiewicz2001efficient}. 

\subsection{Deep learning methods} \label{subsec:DL}

A large number of deep learning (DL) models have been proposed for 3D
point cloud classification, semantic segmentation, object detection, and
registration. PointNet \cite{qi2017pointnet} is among the initial
neural network models for 3D point cloud analysis followed by PointNet++
\cite{qi2017pointnet++}, DGCNN \cite{wang2019dynamic}, etc.  For point
cloud odometry, Nikolai \etal \cite{nicolai2016deep} project 3D point
clouds to panoramic depth images, use a 2D convolutional layers to
extract features, and estimate motion parameters with fully-connected
(FC) layers, leading to an end-to-end network design.  DeepPCO
\cite{wang2019deeppco} uses two separate networks to predict 3
orientation and 3 translation parameters, respectively.  DL methods for
visual odometry have been proposed as well.  Konda \etal
\cite{konda2015learning} use a CNN for visual odometry by relating
depth and motion to velocity change. DeepVO \cite{wang2017deepvo} uses
the recurrent CNN for visual odometry.  ViNet \cite{clark2017vinet}
treats visual odometry as a sequence-to-sequence problem. Other
noteworthy works include Flowdometry \cite{muller2017flowdometry} and
LS-VO \cite{costante2018ls}. 

\subsection{Green Learning and PointHop++} \label{subsec:R-PointHop}

In light of heavy hardware and computational complexity of DL methods, a
green learning methodology targeting mobile and edge applications has
been developed by Kuo and his collaborators recently for several point
cloud processing tasks \cite{kadam2020unsupervised, kadam2021r,
kadam2022pcrp, zhang2020unsupervised, zhang2021gsip,
zhang2020pointhop++, zhang2020pointhop}. These methods follow the
classical pattern recognition pipeline by decomposing a learning task
into the cascade of two individual modules: 1) unsupervised feature
learning and 2) supervised or unsupervised decision learning.  For
feature learning, it uses training data statistics to learn model
parameters, which are the Saab transform filters
\cite{kuo2019interpretable}, at a single stage.  This process can be
repeated for multiple stages. Multi-stage Saab transform filters are
learned in a one-pass feedforward manner.

PointHop++ \cite{zhang2020pointhop++} is a task-agnostic unsupervised
feature learning method for point clouds. It has been successfully
applied to point cloud classification, segmentation and registration.
Local and global point cloud features are learned in an unsupervised,
feedforward and one pass manner in PointHop++. The process can be
summarized below. For every point in the point cloud, its $k$ nearest
neighbors are found and its local 3D space is partitioned into eight
octants centered at the point. Point attributes are constructed by
taking the mean of all point coordinates in each octant. Then, the
channel-wise Saab transform is conducted to learn multi-hop features.
For point cloud classification, features are aggregated and fed to a
classifier. Here, feature aggregation and classification steps are
irrelevant to the odometry task. Only point-wise features are used. 

\section{Proposed GreenPCO Method} \label{sec:method}

\begin{figure*}[th]
\centering
\includegraphics[scale=0.65]{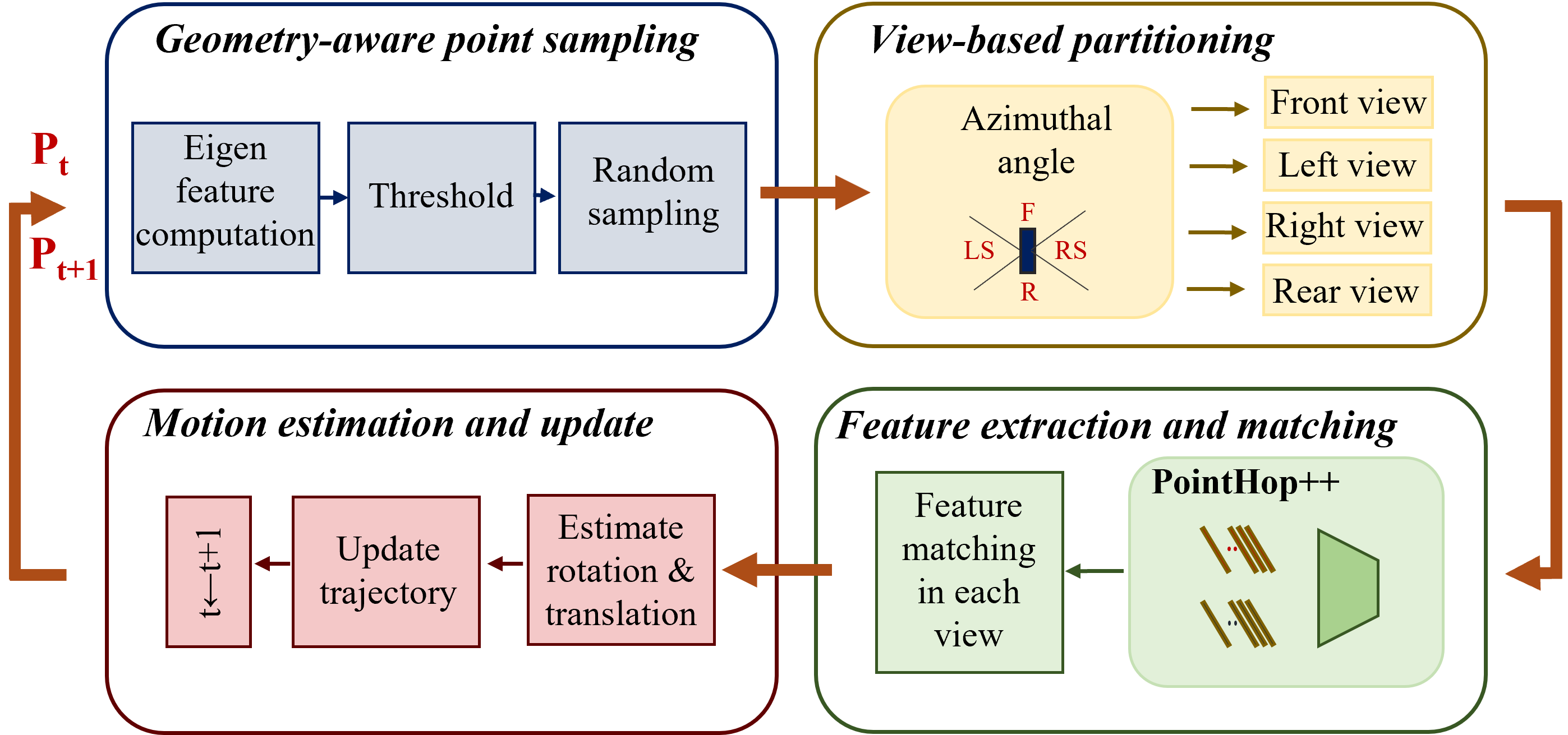}
\caption{An overview of the GreenPCO method.} \label{fig:architecture}
\end{figure*}

The GreenPCO method takes two consecutive point cloud scans at time $t$
and $t+1$ as input and predicts the 6-DOF (Degree of Freedom) rigid
transformation parameters as output. The transformation can be written
in form of $(R, t)$, where $R$ and $t$ denote the rotation matrix and
the translation vector, respectively. Let $x_i$ and $y_i$ be
corresponding points in point cloud scans at $t$ and $t+1$,
respectively.  The objective is to find $R$ and $t$ so as to minimize
the mean squared error
\begin{equation}\label{eq:optimal}
E(R,t)=\frac{1}{N}\sum\limits_{i=1}^{N}\|R \cdot x_i+t-y_i\|^2
\end{equation}
between matching point pairs $(x_i,y_i)$, $i=1, \cdots, N$.  GreenPCO is an
unsupervised learning method since we do not need to create different
rotational and translational sequences for model training. All model
parameters can be learned from raw training sequences.  The system
diagram of GreenPCO is shown in Fig. \ref{fig:architecture}. It consists of
four steps: 1) Geometry-aware point sampling, 2) view-based
partitioning, 3) feature extraction and matching, and 4) motion
estimation.  They are elaborated below. 

\subsection{Geometry-aware point sampling} \label{subsec:GA-sampling}

An outdoor point cloud scan captured by the LiDAR sensor typically
consists of hundreds of thousands of points. However, many points in
outdoor environments are featureless and non-discriminant. It is
desirable and sufficient to select a subset of discriminant points to
build a effective correspondence. To this end, we propose a
geometry-aware point sampling method that selects points that are
spatially spread out and with salient local characteristics. 

\begin{figure*}[th]
\centering
\includegraphics[scale=0.205]{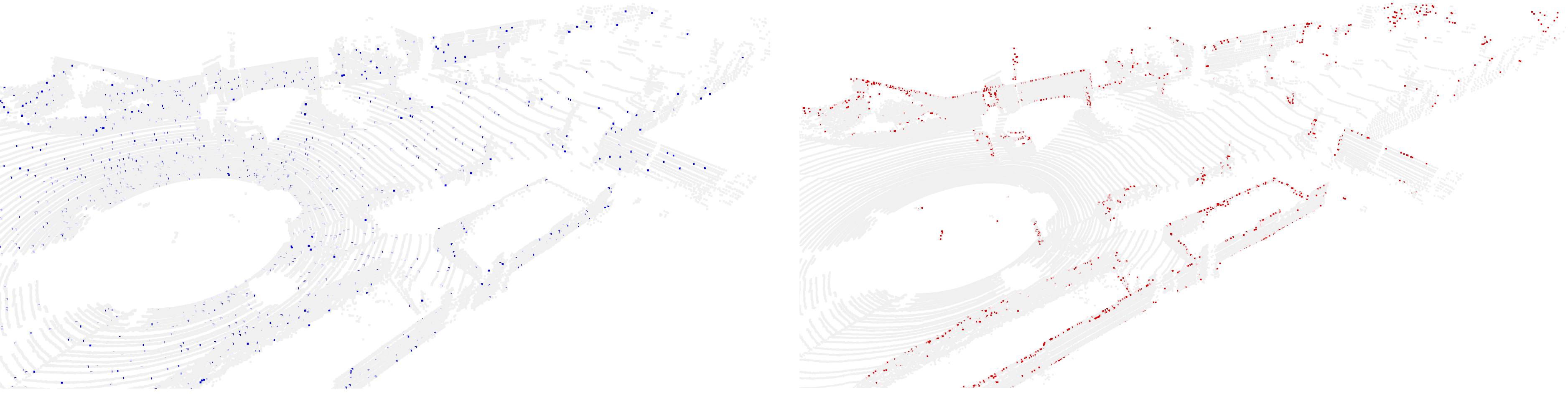}
\caption{Comparison between random sampling (left) and geometry-aware
sampling (right), where sampled points are marked in blue and red,
respectively.} \label{fig:sampling}
\end{figure*}

The local neighborhood of a point, $p$, defines its local property. We
collect $k$ nearest neighbors of $p$ in a local region, find the
covariance matrix of their 3D coordinates, and conduct eigen
decomposition.  The eigenvalues of local PCA can describe the local
characteristics of a point well. For examples, local features such as
linearity, planarity, sphericity, and entropy can be expressed as
functions of the three eigenvalues \cite{hackel2016fast}.  We are
interested in discriminant points (e.g., those from objects like mopeds,
cars, poles, etc.) rather than points from planar surfaces (e.g.,
buildings, roads, walls, etc.) To achieve this goal, we study
distributions of eigenvalues and set appropriate thresholds so as to
discard non-discriminant points in the pre-processing step. In our
implementation, we set thresholds on linearity, planarity, and eigen
entropy.  They are computed as
\begin{eqnarray}
\mbox{Linearity} = \frac{\lambda_1 - \lambda_2}{\lambda_1}&,& \hspace{5mm} 
\mbox{Planarity} =  \frac{\lambda_2 - \lambda_3}{\lambda_1}, \nonumber \\
\mbox{Eigen entropy} &=&  \sum_{i=1}^{3} \lambda_i \cdot \log (\frac{1}{\lambda_i}),
\end{eqnarray}
where $\lambda_1$, $\lambda_2$ and $\lambda_3$ are three eigenvalues of
local PCA, and $\lambda_1 \geq \lambda_2 \geq \lambda_3$. After this
step, we are left with around 4,000 to 5,000 points depending on the
scene. Afterwards, we use random sampling to reduce the point number to
2,048 for further processing in the next step.  Points selected by
geometry-aware point sampling and random sampling are compared in Fig.
\ref{fig:sampling}. We see that the proposed geometry-aware sampling
method preserves more discriminant points from the scene. 

\begin{figure*}[th]
\centering
\includegraphics[scale=0.5]{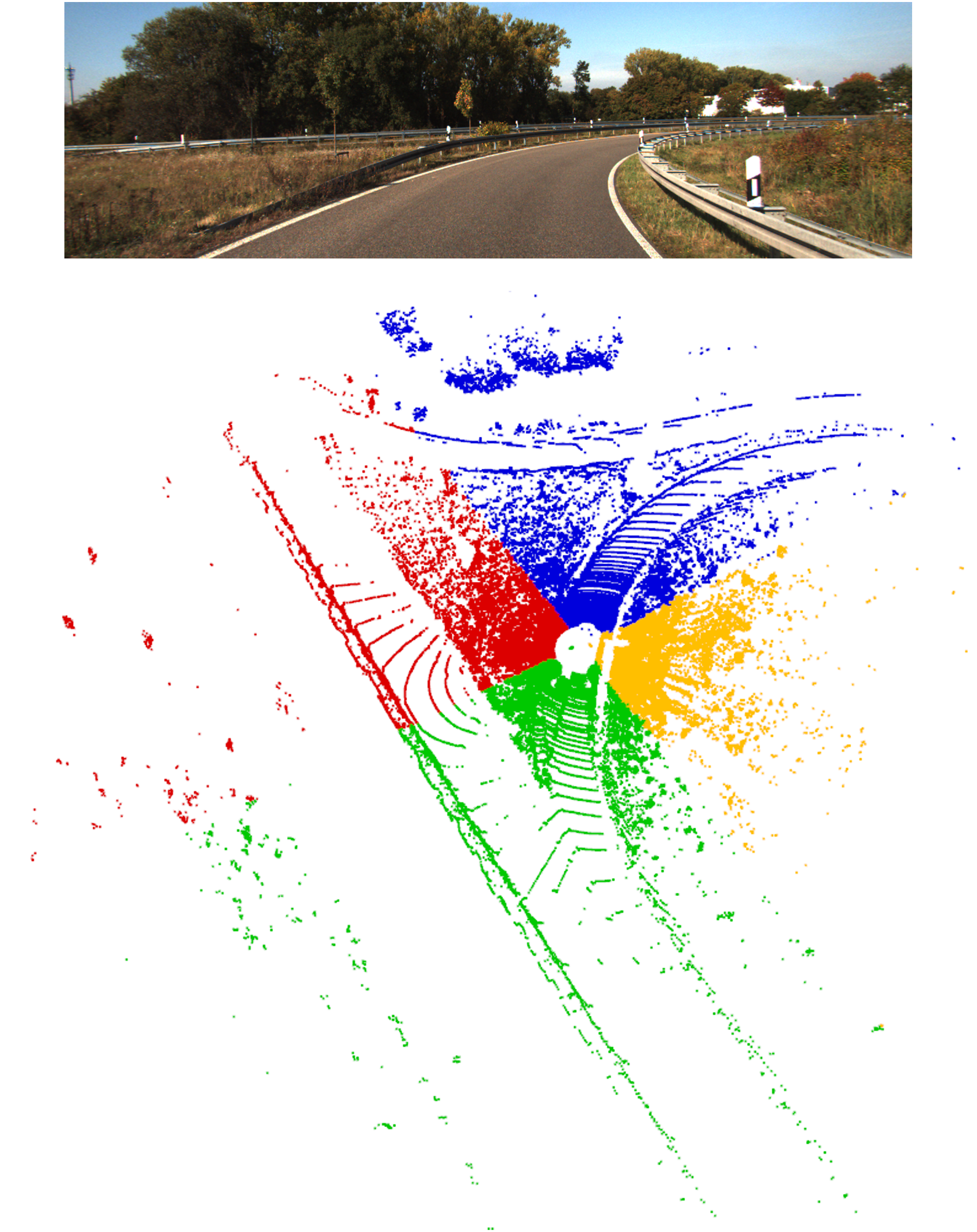}
\caption{View-based partitioning using the azimuthal angle, where the
front, rear, left and right views are highlighted in blue, green, red,
and yellow, respectively.}\label{fig:view-partition}
\end{figure*}

\subsection{View-based partitioning}

The sampled points obtained in the previous step are divided into
disjoint sets based on their 3D coordinates. First, the 3D Cartesian
coordinates are converted to spherical coordinates. Following the
convention of the LiDAR coordinate system setup in the KITTI dataset,
the positive $Z$ direction is along the direction of motion of the
vehicle, while the positive $Y$ direction is vertical. We are interested
in the azimuthal angle $\phi$, which is given by
\begin{equation}
\phi = \arctan (z/x),
\end{equation}
where $z$ and $x$ are point coordinates along the $Z$ and $X$ axes,
respectively. The $\phi$ coordinate of every point gives the position of
the point with respect to the vehicle.  Then, we can define four views
based on $\phi$. Front view contains points with an azimuthal angle
$45^\circ \le \phi \le 135^\circ$.  Rear view contains points with an
azimuthal angle $225^\circ \le \phi \le 315^\circ$.  Right side view
contains points with an azimuthal angle $-45^\circ \le \phi \le
45^\circ$, and left side view contains points with an azimuthal angle
$135^\circ \le \phi \le 225^\circ$.  An example of view-based
partitioning is shown in Fig.  \ref{fig:view-partition}.

Since motion between two consecutive scans is incremental, we can focus
on point matching in the same view.  The proposed partitioning helps in
scenarios when there are similar instances (e.g., persons or equally
spaced poles along the road in different views) in two point cloud
scans.  We also tried to partition points into six disjoint views but
observed no advantage. On the contrary, the probability of correctly
matched points coming from different views increases. Thus, we stick to
the choice of four views. 

\subsection{Feature extraction and point matching}

\begin{figure*}[th]
\centering
\includegraphics[scale=0.275]{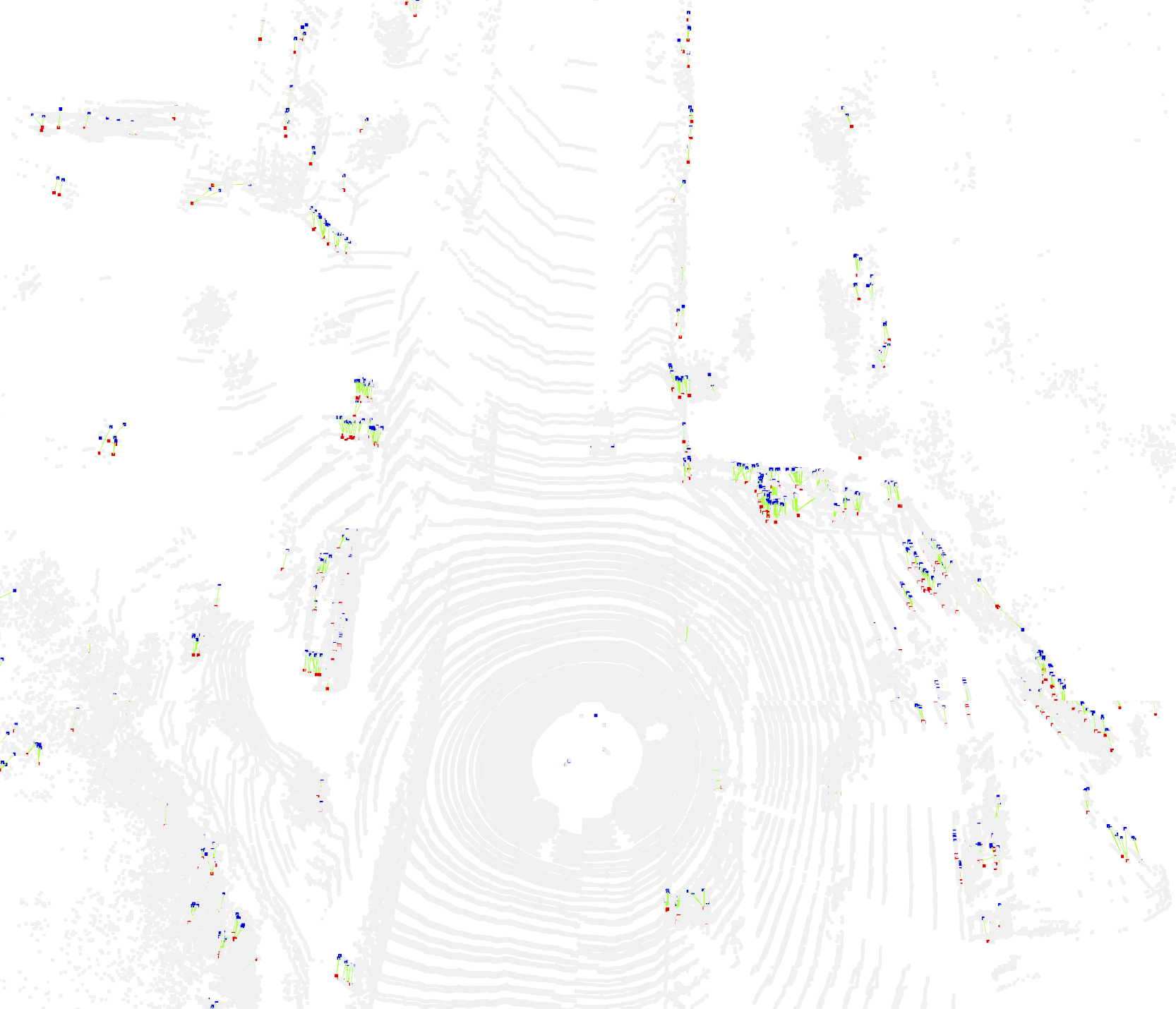}
\caption{Sampled points at time instances $t$ and $t+1$ are marked in blue 
and red, respectively, while point correspondences between two consecutive 
scans are shown in green.} \label{fig:correspondences}
\end{figure*}

The features of all points sampled in Step 1 are extracted using
PointHop++ as described in \cite{zhang2020pointhop++}. Briefly
speaking, the $k$ nearest neighbors of points are retrieved and the 3D
coordinate space is partitioned into 8 octants. The mean of each octant
is calculated and the eight means are concatenated to form the attribute
vector. Then, the channel-wise Saab transform is conducted for dimension
reduction. The attribute construction and dimensionality reduction step
is repeated in the second hop to increase the receptive field. We append
the 3D coordinates with the eigen features \cite{hackel2016fast} to
obtain a rich set of point features.  For point matching in two point
cloud scans, we consider the nearest neighbor search in the feature
space and conduct it in each of the four view groups independently. As a
result, we have four sets of point correspondences - one from each view.
All corresponding pairs are then combined and used to estimate the
rotation and translation between the two scans. A subset of matched
points between two consecutive scans are shown in Fig.
\ref{fig:correspondences}. 

\subsection{Motion estimation and update}

The matched points found in Step 3 are used to estimate the motion
between the two consecutive time instances. Suppose that $(x_i,y_i)$,
$i=1, \cdots, N$, are the pairs of corresponding points. The 6-DOF
motion model can be determined as follows.  First, the mean coordinates
of the corresponding points can be written as
\begin{equation}
\Bar x=\frac{1}{N}\sum\limits_{i=1}^{N} x_i, \quad 
\Bar y=\frac{1}{N}\sum\limits_{i=1}^{N} y_i,
\end{equation}
and the covariance matrix of the corresponding pairs of points can
be computed as
\begin{equation}
K(X,Y)=\sum\limits_{i=1}^{N}(x_i-\Bar{x})(y_i-\Bar{y})^T.
\end{equation}
Next, the $3 \times 3$ covariance matrix can be decomposed via SVD,
\begin{equation}
K(X,Y)= USV^T,
\end{equation}
where $U$ and $V$ are orthogonal matrices of left and right singular
vectors and $S$ is the diagonal matrix of singular values, respectively.
Then, the orientation and translation motion model are given by the 
rotation matrix, $R$, and the translation vector, $t$, in the form of
\begin{equation}
R=VU^T, \mbox{   and   } t=-R\Bar{x}+\Bar{y}.
\end{equation}
The vehicle trajectory is updated with the current predicted pose. The
pose at time $t$, $T^t$, with respect to the pose at time $t-1$,
$T^{t-1}$, is given by $T^t=T^{t-1}[R \ t \ ; \ {\bar{0}}^T \ 1]$. The
pose with respect to the the inital pose $T^0$ can be found accordingly.
The process repeats by considering the next two point cloud scans.
Furthermore, RANSAC \cite{fischler1981random} can be used to improve
the robustness of point matching.

\section{Experimental Results}\label{sec:experiments}

Experiments are conducted on the KITTI Visual Odometry/SLAM benchmark
\cite{Geiger2012CVPR} for performance evaluation. The dataset consists
of 22 sequences in total, out of which the ground truth information is
available for the first 11 sequences. Each sequence is a vehicle
trajectory that consists of from 250 to 2000 time steps. The data at
each time step contains the 3D point cloud scan captured by the LiDAR
scanner, stereo and monocular images. By following point cloud odometry
benchmarking methods such as DeepPCO \cite{wang2019deeppco}, we use the
point cloud data only in the experiments, we use sequences 4 and 10 for
testing and the rest for training. It is observed that there is a strong
correlation between scans of the training data. For this reason, we
uniformly sample 50 point clouds from the nine training sequences to
learn the kernels in channel-wise Saab transform. Furthermore, the
geometry-aware sampling process is used to select discriminant points.
The thresholds for the eigen features during sampling are set to 0.7 for
linearity and planarity and 0.8 for eigen entropy. Points with their
linearity and planarity below the 0.7 threshold and with their eigen
entropy higher than the 0.8 threshold are retained. The neighborhood
size for finding the eigen features is 48 points. Two hops are used in
total. 

\begin{figure*}[th]
\centering
\includegraphics[scale=0.55]{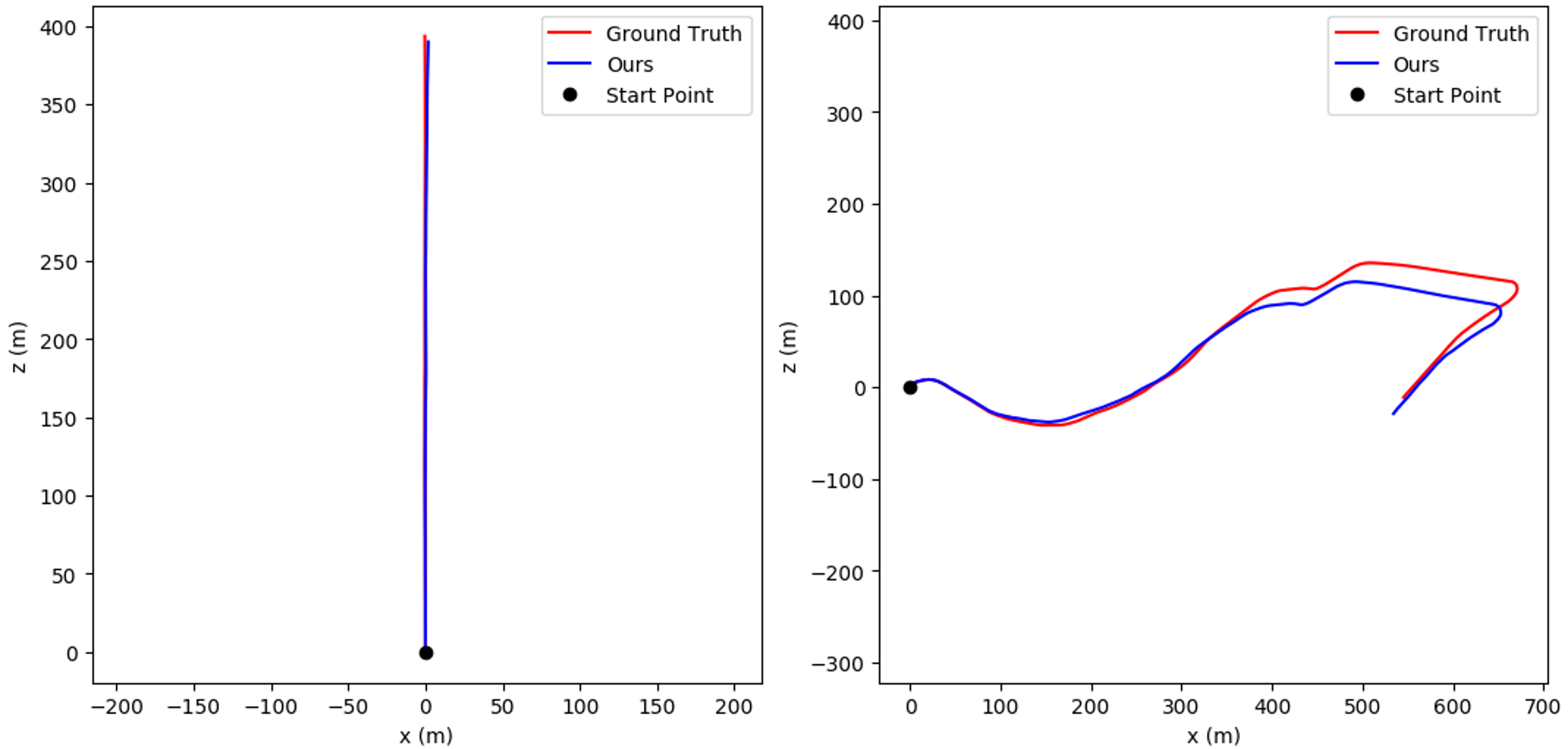}
\caption{Results on sequences 4 (left) and 10 (right) of the
KITTI dataset.} \label{fig:result}
\end{figure*}

The evaluation results for test sequences 4 and 10 are shown in Fig.
\ref{fig:result}. We see that GreenPCO is very effective and its predicted
paths almost overlap with the ground truth ones. Based on the KITTI
evaluation metric, the average sequence translation RMSE is 3.54\% while
the average sequence rotation error is 0.0271 deg/m.  We compare GreenPCO
with five supervised DL methods in Table \ref{tab:Result}.  They are
Two-stream \cite{nicolai2016deep}, DeepVO \cite{wang2017deepvo},
PointNet \cite{qi2017pointnet}, PointGrid \cite{le2018pointgrid} and
DeepPCO \cite{wang2019deeppco}.  The evaluation metrics are the same as
those in DeepPCO \cite{wang2019deeppco}, where relative translation and
rotation errors are considered. Although GreenPCO is an unsupervised
learning method, it outperforms all supervised DL methods in both the
average rotation RMSE and the average translation RMSE. 

\begin{table*}[ht]
\centering
\caption{Performance comparison between GreenPCO and five supervised DL
methods on two test sequences in the KITTI dataset} \label{tab:Result}
\renewcommand\arraystretch{1.3}
\newcommand{\tabincell}[2]{\begin{tabular}{@{}#1@{}}#2\end{tabular}}
\begin{tabular}{c | c c | c c} \hline 
\tabincell{c}{} &\multicolumn{2}{c|}{\textbf{Sequence 4}} 
&\multicolumn{2}{c}{\textbf{Sequence 10}} \\
\tabincell{c}{Method} & \tabincell{c}{Avg. translation \\ RMSE}  &  \tabincell{c}{Avg. rotation \\ RMSE}  
& \tabincell{c}{Avg. translation \\ RMSE} & \tabincell{c}{Avg. rotation \\ RMSE}  \\ \hline 
Two-stream \cite{nicolai2016deep} & 0.0554 & 0.0830  & 0.0870 & 0.1592  \\ \hline
DeepVO \cite{wang2017deepvo}  & 0.2157 & 0.0709  & 0.2153 & 0.3311 \\ \hline
PointNet \cite{qi2017pointnet}  & 0.0946 & 0.0442  & 0.1381 & 0.1360  \\ \hline
PointGrid \cite{le2018pointgrid}  & 0.0550 & 0.0690  & 0.0842 & 0.1523  \\ \hline
DeepPCO \cite{wang2019deeppco} & 0.0263 & 0.0305  & 0.0247 & 0.0659  \\ \hline
GreenPCO (Ours)  & \bf{0.0201} & \bf{0.0212} & \bf{0.0209} & \bf{0.0628}   \\ \hline
\end{tabular}
\end{table*}

We conduct an ablation study on GreenPCO to see the contributions of
each component and summarize the results in Table
\ref{tab:Ablation-study}.  First, we compare geometry-aware, random, and
farthest point sampling methods.  For each sampling method, we set the
number of sampled points to 1024, 2048, and 4096 points. Geometry-aware
sampling consistently outperforms the other two. The errors of random
sampling are worst while errors of farthest point sampling are also
large.  There is no advantage of using 4096 points over 2048 points for
any sampling method. Thus, we set the input point number to 2048. Next,
we justify the inclusion of view-based partitioning and eigen features.
Errors are consistently lower when view-based partitioning is adopted.
This shows that view-based partitioning makes point matching more
robust.  Furthermore, it reduces the search space for time efficiency.
Finally, when eigen features are omitted from the feature construction
process, the performance drops sharply. 

\begin{table*}[httb]
\centering
\caption{Ablation study on KITTI dataset} \label{tab:Ablation-study}
\newcommand{\tabincell}[2]{\begin{tabular}{@{}#1@{}}#2\end{tabular}}
\begin{tabular}{c c c c c c} \hline 
\tabincell{c}{Sampling Method} & \tabincell{c}{Number of \\ sampled points}  &  \tabincell{c}{View-based \\ partitioning}
& \tabincell{c}{Eigen \\ features} & \tabincell{c}{Translation \\ error (\%)} & \tabincell{c}{Rotation error \\  (deg/m)} \\ \hline 
\multirow{3}{*}{Random} & 1024 & \checkmark  & \checkmark  & 32.42 & 0.1120  \\ 
& 2048 & \checkmark & \checkmark  & 32.20 & 0.1179 \\ 
 & 4096 & \checkmark & \checkmark  & 31.47 & 0.1004  \\ \hline
\multirow{3}{*}{Farthest point} & 1024 & \checkmark & \checkmark  & 28.79 & 0.1065 \\ 
 & 2048 & \checkmark & \checkmark  & 28.11 & 0.0971 \\ 
  & 4096 & \checkmark & \checkmark  & 28.13 & 0.0941   \\ \hline
\multirow{6}{*}{Geometry-aware} & 1024 & \checkmark  & \checkmark  & 3.71 & 0.0345 \\ 
  & 2048 & \checkmark & \checkmark  & \bf{3.54} & \bf{0.0271} \\ 
  & 4096 & \checkmark  & \checkmark  & \bf{3.54} & 0.0308  \\
  & 2048 &   &  & 10.41 & 0.0415  \\ 
 & 2048 &  & \checkmark  & 4.89 & 0.0289  \\ 
  & 2048 & \checkmark  &  & 9.81 & 0.0401  \\ \hline
\end{tabular}
\end{table*}

Since most of the model-free methods such as LOAM \cite{zhang2014loam}
and V-LOAM \cite{zhang2015visual} already offer state-of-the-art
results, spending a large amount of time on model training (like deep
learning) is not an optimal choice for this problem. GreenPCO is more
favorable in this sense. Its training time is only 10 minutes on Intel
Xeon CPU. 

The rapid training time of GreenPCO is attributed to two reasons. First,
PointHop++ model training demands sampled points to learn filter
parameters. Other steps in the pipeline such as view-based partitioning,
point matching, and motion estimation are only required during testing.
Second, we use an extremely small training set since there is a high
correlation between consecutive point cloud scans in different sequences
due to incremental vehicular motion. Hence, skipping several scans and
selecting only distant scenes should be sufficient to capture different
scenes within the training data. It is observed that the use of 50 point
cloud scans offers performance similar to that using the entire training
dataset which comprises of 8500 scans. This corresponds to approximately
0.6\% of the training data. 

In our implementation, the 50 point cloud scans are uniformly sampled
from the entire training dataset so as to represent diverse scenes.
Table \ref{tab:train_time} summarizes the test performance for four
different percents of training data. As shown in the table, the
translation and rotation errors are unaffected and there is no advantage
of using the entire training data. There is a steady decrease in the
training time as the size of the training data is reduced.  The model
size is 75kB, which is independent of the training data used. A small
model size and less training time make GreenPCO a lightweight solution
for point cloud odometry. 

\begin{table}[ht]
\centering
\caption{The effect of different amounts of training data} \label{tab:train_time}
\renewcommand\arraystretch{1.3}
\newcommand{\tabincell}[2]{\begin{tabular}{@{}#1@{}}#2\end{tabular}}
\begin{tabular}{c c c c c} \hline
\tabincell{c}{Training data \\ used (\%)} & \tabincell{c}{Training \\  time (hours)}  &  \tabincell{c}{Model size}  
& \tabincell{c}{Translation \\ error (\%)} & \tabincell{c}{Rotation error \\ (deg/m)}  \\ \hline
100 & 1.2 & 75kB & 3.54 & 0.0268 \\ \hline
50  & 0.8 & 75kB & 3.53 & 0.0272 \\ \hline
25  & 0.6 & 75kB & 3.54 & 0.0271 \\ \hline
10  & 0.5 & 75kB & 3.54 & 0.0271 \\ \hline
0.6  & 0.17 & 75kB & 3.54 & 0.0271 \\ \hline
\end{tabular}
\end{table}

\section{Conclusion and Future Work}\label{sec:conclusion}

An unsupervised lightweight learning method for point cloud odometry,
called GreenPCO, was proposed in this work. GreenPCO takes consecutive
point cloud scans captured by the LiDAR sensor on the vehicle and
estimates the 6-DOF motion of the vehicle incrementally. It first
selects a small set of discriminant points using geometry-aware
sampling. Then, the sampled points are divided into four disjoint sets
based on the azimuthal angle.  The point features are extracted using
the PointHop++ method and matching points are found by searching
neighbors in the feature space. Finally, the corresponding points are
used to estimate the rotation and translation between the two positions.
The same process is repeated for subsequent scans. GreenPCO gives
accurate vehicle trajectories when evaluated on the LiDAR scans from the
KITTI dataset. It outperforms all supervised DL-based benchmarking
methods with much less training data. 

As a future extension, it is worthwhile to analyze failure cases of the
model-free methods based on scan matching and see whether replacing
their handcrafted features with our learned features helps overcome
their drawbacks.  This opens a new research direction in exploiting
traditional methods with unsupervised feature learning based on data
statistics.  Also, we applied GreenPCO to an outdoor PCO in this work.
We expect that the same technique can be extended to indoor and/or other
restricted environments with minor adjustments. The generalization to
mobile robots or other moving agents should be valuable in
real-world applications.

\bibliographystyle{unsrt}
\bibliography{refs}

\end{document}